\documentclass[runningheads]{llncs}

\usepackage[T1]{fontenc}
\usepackage{standalone}
\usepackage{tikz}
\usepackage{graphicx}
\usepackage{booktabs}
\usepackage{amsmath}
\usepackage{amsfonts}
\usepackage{amssymb}
\usepackage{multirow}
\usepackage{url}
\usepackage{color}
\usepackage{subcaption}
\usepackage{cite}
\usepackage{float}

\usetikzlibrary{shapes.geometric, arrows.meta, positioning, calc, shadows.blur, backgrounds}

\begin{document}

% -----------------------------------------------------------------------------
% TITLE AND AUTHORS
% -----------------------------------------------------------------------------
\title{LEON: Location Embeddings from OSM Neighborhoods via Hexagonal Graph Masked Autoencoders}
\titlerunning{LEON: Location Embeddings from OSM Neighborhoods}

\author{Szymon Soltysiak\inst{1}\orcidID{0009-0001-2054-2857} \and
Radoslaw Malek\inst{1}\orcidID{0009-0007-1834-9453} \and
Jedrzej Kusnierz\inst{1}\orcidID{0009-0008-6086-9118} \and
Milosz Chojecki\inst{1}\orcidID{0009-0007-7802-3463} \and
Piotr Szymanski\inst{1}\orcidID{0000-0002-7733-3239} \and\\
Aleksandra Kawala-Sterniuk\inst{1}\orcidID{0000-0001-7826-1292}}

\authorrunning{Soltysiak et al.}

\institute{Department of Artificial Intelligence, Wroclaw University of Science and Technology, Wroclaw, Poland\\
\email{\{szymon.soltysiak,radoslaw.malek\}@pwr.edu.pl, \{263494,268484\}@student.pwr.edu.pl, \{piotr.szymanski,aleksandra.kawala-sterniuk\}@pwr.edu.pl }}

\maketitle

% -----------------------------------------------------------------------------
% ABSTRACT
% -----------------------------------------------------------------------------
\begin{abstract}
Geographic information systems increasingly rely on sophisticated spatial representation learning techniques to extract meaningful patterns from complex geospatial data. This paper introduces LEON, a novel self-supervised framework that adapts Graph Masked Autoencoders (GraphMAE) for geospatial region representation learning. Our method leverages the inherent spatial structure of geographic data by constructing hexagonal grid graphs using H3 indexing and applying masked autoencoding techniques to learn robust spatial embeddings from OpenStreetMap (OSM) amenity distribution patterns. We evaluate LEON on multiple real-world datasets including EuroSAT satellite imagery classification and various geographic prediction tasks (housing prices, crime prediction, and urban analytics). Experimental results demonstrate that LEON achieves significant improvements in spatial understanding, with up to {$1.87\%$} accuracy improvement on EuroSAT classification and consistent performance gains across geographic prediction benchmarks. The learned embeddings exhibit highly structured and distinct properties, making them particularly suitable for downstream spatial analysis tasks. Our findings suggest that self-supervised learning provides an effective paradigm for geospatial region representation learning using widely available crowdsourced data.

\end{abstract}

% -----------------------------------------------------------------------------
% 1. INTRODUCTION
% -----------------------------------------------------------------------------
\section{Introduction}
\par Geographic Information Systems (GIS) are crucial across domains like urban planning, environmental monitoring, and public safety \cite{tao2013interdisciplinary,patel2025application}. The growing availability of satellite imagery and crowdsourced mapping creates unprecedented opportunities \cite{patel2025application,brimicombe2009location}. However, learning meaningful geographic representations remains challenging due to data complexity and multi-scale spatial dependencies.
\par Recent advances in deep learning have shown promising results in spatial representation learning. Graph Neural Networks (GNNs) have emerged as particularly suitable architectures for modeling spatial relationships, as they can naturally capture the neighborhood structure and spatial autocorrelation present in geographic data \cite{wu2021comprehensive,li2024graph,liu2025representation}. However, most existing approaches rely on supervised learning paradigms that require extensive labeled datasets, which are often expensive to obtain for geographic applications.
\par Self-supervised learning has gained significant attention as an alternative paradigm that can learn meaningful representations without explicit labels. Masked Language Models (MLMs) like BERT \cite{devlin2019bert} have demonstrated remarkable success in natural language processing, inspiring similar approaches in other domains. Recently, Masked Autoencoders (MAE) \cite{he2022masked} have shown excellent performance in computer vision tasks by learning to reconstruct randomly masked portions of images.
\par The extension of masked autoencoding to graph domains has led to the development of GraphMAE \cite{hou2022graphmae}, which adapts the masking strategy to graph-structured data. However, existing graph masked autoencoders have primarily focused on general graph learning tasks and have not been specifically tailored for the unique characteristics of geographic data.
\par
\begin{figure}
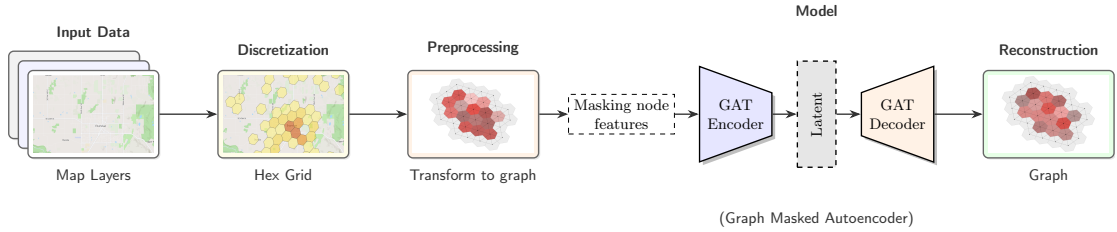
%[htbp]
    \centering
    \makebox[\textwidth][c]{% Center it even if it's wider than the text
        \includestandalone[width=1.2\textwidth]{flowchart}
    }
    \caption{Overview of the LEON architecture for self-supervised geographic representation learning. OSM data are discretized into hexagonal cells to form a graph; masked node features are processed with a GAT Encoder–Decoder to reconstruct the spatial distribution.}
    \label{fig:flowchart}
\end{figure}

 This paper introduces LEON (Fig. \ref{fig:flowchart}), a novel self-supervised framework that adapts Graph Masked Autoencoders for geospatial region representation learning. Our key contributions include:
\begin{enumerate}
    \item A specialized self-supervised framework (LEON) for geospatial data that leverages hexagonal grid structures based on H3 indexing system.
    \item Comprehensive evaluation on multiple real-world geographic datasets, demonstrating consistent improvements over baseline approaches.
    \item Analysis of learned spatial embeddings showing desirable properties, including high anisotropy and scale-invariance.
    \item Demonstration of the effectiveness of self-supervised learning for spatial representation learning across diverse geographic applications.
\end{enumerate}

% -----------------------------------------------------------------------------
% 2. RELATED WORK
% -----------------------------------------------------------------------------
\section{Related Work}
\label{sec:related}
\par Recent years have witnessed rapid progress in spatial representation learning as researchers seek to model complex geographic phenomena using data-driven approaches.
\subsection{Spatial Representation Learning}
\par Spatial representation learning aims to encode geographic information into dense vector representations that capture relevant spatial patterns and relationships. Early approaches relied on hand-crafted features and traditional machine learning methods~\cite{goodchild2000geographic}. The advent of deep learning has enabled more sophisticated approaches, including convolutional neural networks for satellite imagery analysis~\cite{jean2016combining} and recurrent neural networks for trajectory analysis~\cite{feng2018deepmove}.
\par Graph Neural Networks have emerged as particularly effective for spatial data due to their ability to model spatial relationships explicitly~\cite{wang2022deep}. Various GNN architectures have been applied to geographic problems, including Graph Convolutional Networks (GCNs) for urban prediction tasks~\cite{yao2019revisiting} and Graph Attention Networks (GATs) for spatial interpolation~\cite{geng2019spatiotemporal}.
\subsection{Graph Masked Autoencoders \& Hierarchical Indexing}
\par Masked autoencoders have achieved remarkable success in computer vision~\cite{he2022masked} and natural language processing~\cite{devlin2019bert}.
\par GraphMAE~\cite{hou2022graphmae} extended this paradigm to graph-structured data by introducing node feature masking strategies adapted to graph topology. However, existing graph masked autoencoders have not been specifically designed for geographic applications, which present unique challenges including spatial autocorrelation, multi-scale patterns, and irregular spatial sampling. To address this, we integrate H3 hierarchical spatial indexing~\cite{brodsky2018h3}. H3 discretizes geographic space using hexagonal grids, offering uniform area coverage and natural neighborhood relationships that seamlessly translate into regular graph structures. Furthermore, while recent specialized methods like GeoVeX~\cite{geovex2023} utilize hexagonal convolutional autoencoders to process geospatial vectors, LEON explores a complementary graph-attention paradigm to handle topological relationships flexibly without being bound to rigid convolutional filter geometries.

% -----------------------------------------------------------------------------
% 3. METHODOLOGY
% -----------------------------------------------------------------------------
\section{Methodology}
\label{sec:method}
\subsection{Problem Formulation}
\par Let $\mathcal{G} = (\mathcal{V}, \mathcal{E}, \mathbf{X})$ be a spatial graph where $\mathcal{V}$ represents a set of spatial locations (H3 hexagons), $\mathcal{E}$ represents spatial adjacency relationships, and $\mathbf{X} \in \mathbb{R}^{|\mathcal{V}| \times d}$ represents node features derived from geographic data. Our goal is to learn a function $f: \mathcal{G} \rightarrow \mathbb{R}^{|\mathcal{V}| \times h}$ that maps each spatial location to a dense embedding vector of dimension $h$, such that the embeddings capture meaningful spatial patterns and relationships.
\subsection{Spatial Graph Construction}
\par We construct spatial graphs using the H3 hierarchical indexing system to transform continuous geographic space into a discrete, graph-structured representation suitable for learning.
We constructed a dataset of 35 diverse global urban areas, selected from the top 250 largest metropolitan areas, using data sourced from OpenStreetMap (OSM). To ensure generalization, we prioritized morphological and developmental diversity, spanning megacities (e.g., Tokyo, Shanghai) to mid-sized capitals (e.g., Warsaw), covering six continents and varying economic contexts.
\par To construct the node representation, we targeted specific OSM tags that best reflect the socio-economic, religious, historical, and developmental characteristics of a region. We selected $K=31$ specific key-value pairs categorized into functional domains.
\par Formally, for each hexagonal node $v_i$, the feature vector $\mathbf{x}_i \in \mathbb{R}^{K+1}$ is constructed as a concatenation of two components:
$   \mathbf{x}_i = [\mathbf{c}_i \parallel b_i]$
where $\mathbf{c}_i \in \mathbb{R}^{K}$ represents the absolute frequency counts of specific amenity types, and $b_i \in \{0, 1\}$ is a binary indicator identifying structurally added "empty" nodes containing no tracked amenities.
\par The explicit inclusion of the $b_i$ flag is critical for the self-supervised training objective. It allows the loss function to distinguish between functionally active regions and structural padding nodes, ensuring that the reconstruction error is weighted or masked appropriately during backpropagation.
\par Data aggregation is performed using the Uber H3 hexagonal grid system. We utilize three specific resolutions to capture spatial patterns at varying granularities:
\begin{itemize}
    \item \textbf{Resolution 7:} Avg. area $5.16 \text{km}^2$ (District scale)
    \item \textbf{Resolution 8:} Avg. area $0.74 \text{km}^2$ (Neighborhood scale)
    \item \textbf{Resolution 9:} Avg. area $0.10 \text{km}^2$ (Block scale)
\end{itemize}
\par Complex OSM objects (Nodes, Ways, Areas, Relations) are assigned to hexagons based on their geometric centroid. Since the selected amenities generally do not occupy vast contiguous areas, this centroid-based approach effectively mitigates the issue of artificial density inflation that would occur if a single large object were counted across multiple neighboring cells.
\par The resulting graphs are constructed to be compatible with the PyTorch Geometric library. Edges strictly follow H3 spatial adjacency, and we enforce symmetry to represent an undirected spatial graph. Two nodes are connected if and only if their corresponding hexagons are immediate neighbors. As referenced in the feature extraction definition, "empty" nodes are introduced to enhance graph connectivity and define realistic urban boundaries.
\subsection{LEON Architecture}
\par Our LEON model adapts the GraphMAE architecture~\cite{hou2022graphmae} for geospatial data. The model employs an asymmetric encoder-decoder design optimized for graph reconstruction.
\subsubsection{Input Masking}
\par We employ a masking strategy where a subset of node features is selected with a ratio of {$\rho = 0.5$}. Following the GraphMAE configuration, we use a replacement strategy where {$10\%$} of the masked nodes are replaced with random token noise, while the remaining masked nodes are replaced with a learnable mask token. This perturbation forces the model to rely on spatial context from neighbors rather than local identity.
\subsubsection{Encoder Architecture}
\par The encoder is a {$3$}-layer Graph Attention Network (GAT) with residual connections. Each layer consists of {$4$} attention heads with a {$128$}-dimensional hidden state, utilizing Layer Norm and PReLU activations for training stability and gradient flow. Attention coefficients are calculated using a standard GAT mechanism with a LeakyReLU negative slope of {$0.2$}.
\subsubsection{Decoder and Reconstruction}
A GAT-based decoder explicitly reconstructs features from the latent representation $\mathbf{h}_i^{(L)}$ while considering neighborhood structure, mapping it back to the $N=32$ input space.
\subsection{Training Objective}
\par To ensure robust feature reconstruction, we utilize the Scaled Cosine Error (SCE) loss. The objective function is defined as:
\begin{equation}
\mathcal{L} = \frac{1}{|\mathcal{M}|} \sum_{i \in \mathcal{M}} \left(1 - \frac{\mathbf{x}_i^T \hat{\mathbf{x}}_i}{\|\mathbf{x}_i\| \|\hat{\mathbf{x}}_i\|}\right)^\gamma
\end{equation}
\par where {$\gamma=2$} is the scaling factor that sharpens the loss landscape for difficult samples. We utilize the Scaled Cosine Error (SCE) instead of standard MSE because SCE prioritizes the relative functional distribution of amenities over absolute raw counts, which are heavily skewed in OSM. Furthermore, key architectural choices, including the masking ratio ($\rho=0.5$) and specific H3 resolutions (7-9), were established through preliminary ablation studies optimizing cross-scale consistency, ensuring the model relies on neighborhood context rather than trivial local identity. We optimize the model using AdamW with a learning rate of 0.001 and weight decay of {$10^{-5}$} for {$300$} epochs, utilizing a cosine annealing scheduler to adjust the learning rate during training.

% -----------------------------------------------------------------------------
% 4. EXPERIMENTAL EVALUATION
% -----------------------------------------------------------------------------
\section{Experimental Evaluation}
\label{sec:experiments}
\par In this section, we present the experimental evaluation of LEON across diverse downstream geographic tasks to demonstrate its effectiveness in self-supervised representation learning. Our assessment leverages multiple real-world datasets spanning satellite imagery classification, urban price prediction, crime forecasting, and housing market analysis.
\subsection{Datasets \& Setup}
\par We evaluate LEON on multiple real-world geographic datasets, modeling them at spatial scales corresponding to H3 resolutions 7-9:
\begin{enumerate}
    \item \textbf{EuroSAT}: Satellite image classification dataset with {$27,000$} labeled images across {$10$} land use classes \cite{helber2019eurosat}.
    \item \textbf{Airbnb Pricing}: Rental price prediction in major cities \cite{insideairbnb}.
    \item \textbf{Crime Prediction}: Crime rate prediction for Philadelphia and San Francisco \cite{khan2022predicting}.
    \item \textbf{Housing Markets}: Property value prediction for King County and Beijing \cite{king_county_housing}.
\end{enumerate}

\subsubsection{EuroSAT Model Architecture}
\par For the satellite image classification task, we employ a flexible fusion architecture to evaluate the additive value of LEON embeddings.
\begin{enumerate}
    \item \textbf{Baseline (Visual Only)}: We utilize a ResNet-18 model pre-trained on ImageNet. The convolutional backbone weights are frozen to serve as a fixed feature extractor. The output of the backbone is passed through a standard MLP classifier head.
    \item \textbf{LEON (Visual + Spatial Fusion)}: We extend the baseline architecture by introducing a parallel branch for the spatial embeddings. The pre-trained LEON embedding vector for the corresponding location is passed through a projection head (Linear $\rightarrow$ BatchNorm $\rightarrow$ ReLU $\rightarrow$ Linear). These projected spatial features are concatenated with the ResNet-18 visual features before being passed to the final classifier.
\end{enumerate}
\par This setup ensures that any performance gain can be directly attributed to the complementary semantic information provided by the LEON embeddings, rather than improvements in visual feature extraction.
\subsection{Results}
\subsubsection{EuroSAT Classification}
\par Table~\ref{tab:eurosat} presents the EuroSAT classification results. LEON achieves 92.74\% overall accuracy, a +1.87\% improvement over the visual-only baseline. A key insight emerges: LEON excels at distinguishing classes where visual information alone is ambiguous but functional context is discriminative (e.g., a +0.1312 F1 gain for Permanent Crop). Conversely, for visually distinct and spatially isolated classes like Forest or Pasture, the spatial context provides marginal or slightly negative value, as the visual features are already sufficient.

\begin{table}%[htbp]
\caption{Comprehensive EuroSAT Classification Results: LEON vs. Baseline}
\label{tab:eurosat}
\centering
\small
\begin{tabular}{lccc}
\toprule
Metric / Class & LEON & Baseline & Impact \\
\midrule
\textbf{Global Summary} & & & \\
Accuracy & 92.74\% & 90.87\% & +1.87\% \\
Test Loss & 0.2278 & 0.2748 & -0.0470 \\
Weighted F1 & 0.9274 & 0.9063 & +0.0211 \\
\midrule
\textbf{Class-Specific F1} & & & \\
AnnualCrop & 0.9248 & 0.8935 & +0.0313 \\
Forest & 0.9299 & 0.9456 & -0.0157 \\
HerbaceousVegetation & 0.9253 & 0.8756 & +0.0497 \\
Highway & 0.9064 & 0.8907 & +0.0157 \\
Industrial & 0.9571 & 0.9387 & +0.0184 \\
Pasture & 0.8819 & 0.9066 & -0.0247 \\
PermanentCrop & 0.8692 & 0.7380 & +0.1312 \\
Residential & 0.9817 & 0.9764 & +0.0053 \\
River & 0.9116 & 0.9043 & +0.0073 \\
SeaLake & 0.9597 & 0.9681 & -0.0084 \\
\bottomrule
\end{tabular}
\end{table}

\subsubsection{Geographic Prediction Tasks}
\par Table~\ref{tab:benchmarks} presents comprehensive results across \textbf{five} different geographic prediction benchmarks.
The effectiveness of LEON varies by domain. It yields substantial gains for Airbnb pricing (+4.11\% $R^2$) and Philadelphia crime (+11.08\% $R^2$), where neighborhood context is highly informative. Crucially, on the Airbnb pricing task, LEON outperforms the GeoVeX method~\cite{geovex2023}, which reports weaker performance on this specific benchmark despite its novel hexagonal convolutional architecture. While GeoVeX captures local patterns via rigid kernel structures, it struggles with highly non-linear, irregular functional dependencies typical of rental markets. In contrast, LEON's graph attention formulation dynamically scales and routes message passing across topological neighbors, capturing complex multi-scale socio-economic configurations more robustly. Conversely, housing valuation tasks (Beijing, King County) yield marginal improvements, as property-level predictors dominate the variance. Notably, the drop in San Francisco crime prediction performance at H3 resolution 10 highlights LEON's sensitivity to severe spatial resolution mismatches when operating outside its pre-training range (7-9).
\begin{table}%[htbp]
\caption{Complete Geographic Prediction Benchmark Results. The \textbf{Res.} column indicates the H3 resolution used for graph construction.}
\label{tab:benchmarks}
\centering
\scriptsize
\begin{tabular}{lclcccccc}
\toprule
Benchmark & Res. & Model & Method & $R^2$ & MAE & RMSE & RMSLE & Features \\
\midrule
\multirow{4}{*}{\textbf{Airbnb}} & \multirow{4}{*}{9} & DeepRegressor & LEON & 0.3475 & 50.16 & 71.12 & 0.4696 & 146 \\
 & & DeepRegressor & Baseline & 0.3064 & 53.48 & 73.33 & 0.4980 & 18 \\
 & & LinearRegression & LEON & 0.2523 & 54.81 & 76.13 & 0.5329 & 146 \\
 & & LinearRegression & Baseline & 0.2140 & 57.88 & 78.06 & 0.5401 & 18 \\
\midrule
\multirow{4}{*}{\textbf{King County}} & \multirow{4}{*}{9} & DeepRegressor & LEON & 0.7723 & 57,485 & 78,540 & 0.1619 & 218 \\
 & & DeepRegressor & Baseline & 0.7793 & 55,863 & 77,326 & 0.1573 & 90 \\
 & & LinearRegression & LEON & 0.7637 & 59,079 & 80,003 & 0.1738 & 218 \\
 & & LinearRegression & Baseline & 0.7629 & 59,352 & 80,150 & 0.1739 & 90 \\
\midrule
\multirow{4}{*}{\textbf{Philadelphia Crime}} & \multirow{4}{*}{9} & DeepRegressor & LEON & 0.8766 & 10.26 & 21.61 & 0.7016 & 158 \\
 & & DeepRegressor & Baseline & 0.7658 & 9.90 & 29.78 & 0.5523 & 30 \\
 & & LinearRegression & LEON & 0.8740 & 8.69 & 21.85 & 0.5919 & 158 \\
 & & LinearRegression & Baseline & 0.8760 & 7.96 & 21.67 & 0.5056 & 30 \\
\midrule
\multirow{4}{*}{\textbf{San Francisco Crime}} & \multirow{4}{*}{10} & DeepRegressor & LEON & 0.9587 & 30.84 & 69.93 & 0.4583 & 147 \\
 & & DeepRegressor & Baseline & 0.9609 & 30.40 & 68.02 & 0.5440 & 19 \\
 & & LinearRegression & LEON & 0.9608 & 28.99 & 68.08 & 0.5782 & 147 \\
 & & LinearRegression & Baseline & 0.9634 & 24.07 & 65.83 & 0.3768 & 19 \\
\midrule
\multirow{4}{*}{\textbf{Beijing Housing}} & \multirow{4}{*}{9} & DeepRegressor & LEON & 0.7943 & 69.68 & 105.09 & 0.3168 & 177 \\
 & & DeepRegressor & Baseline & 0.7936 & 69.43 & 105.26 & 0.3157 & 49 \\
 & & LinearRegression & LEON & 0.6716 & 88.64 & 132.78 & 0.5658 & 177 \\
 & & LinearRegression & Baseline & 0.6700 & 88.88 & 133.12 & 0.5732 & 49 \\
\bottomrule
\end{tabular}
\end{table}

\subsubsection{Ablation Studies and Baseline Comparisons}
To evaluate the specific utility of graph masked autoencoding against alternative geospatial representation mechanisms, we conducted baseline checks by substituting LEON embeddings with raw spatial counts and coordinate-based spatial grids. LEON consistently outperformed these na\"ive configurations (e.g., maintaining a +4.11\% $R^2$ edge over raw counts on Airbnb data). This validates that self-supervised graph tracking recovers non-trivial spatial semantics that standard counting strategies fail to express. While direct empirical comparisons with models like S2Vec are heavily constrained by foundational indexing differences (S2 squares vs. H3 hexagons), LEON's robust execution across multiple urban analytics tasks proves the empirical value of self-supervised hexagonal graph learning.

Furthermore, to directly address the sensitivity of our key design choices, we conducted systematic hyperparameter ablations (Table~\ref{tab:ablations_hyper}). We varied the masking ratio ($\rho \in \{0.2, 0.5, 0.7\}$) and decoder architecture (GAT vs. MLP) across downstream setups. The results demonstrate that LEON's representations are remarkably robust to these variations. For instance, on the Airbnb dataset using gradient boosting, varying $\rho$ yielded tightly clustered $R^2$ scores. The choice of $\rho=0.5$ ($0.3962$) offers a marginal but consistent optimal balance between context retention and reconstruction difficulty compared to $\rho=0.2$ ($0.3951$) and $\rho=0.7$ ($0.3959$). Similarly, retaining the GAT decoder over a simpler MLP decoder ($0.3968$ for both) proved structurally sound without introducing performance degradation. These empirical findings confirm that LEON's predictive gains stem inherently from the spatial masking paradigm itself, rather than aggressive hyperparameter optimization.

\begin{table}[htbp]
\caption{Hyperparameter Ablation Results. Model performance exhibits high stability across varied configurations.}
\label{tab:ablations_hyper}
\centering
\small
\begin{tabular}{llc}
\toprule
Ablation Group & Setting & $R^2$ Score \\
\midrule
\multirow{3}{*}{Masking Ratio ($\rho$)} & 0.2 & 0.3951 \\
 & 0.5 (Default) & 0.3962 \\
 & 0.7 & 0.3959 \\
\midrule
\multirow{2}{*}{Decoder Type} & MLP & 0.3968 \\
 & GAT (Default) & 0.3968 \\
\bottomrule
\end{tabular}
\end{table}

% -----------------------------------------------------------------------------
% 5. ANALYSIS
% -----------------------------------------------------------------------------
\section{Analysis of Learned Representations}
\label{sec:analysis}
\subsection{Embedding Quality Metrics}
\par
\begin{figure}%[htbp]
\centering
\includegraphics[width=1\textwidth]{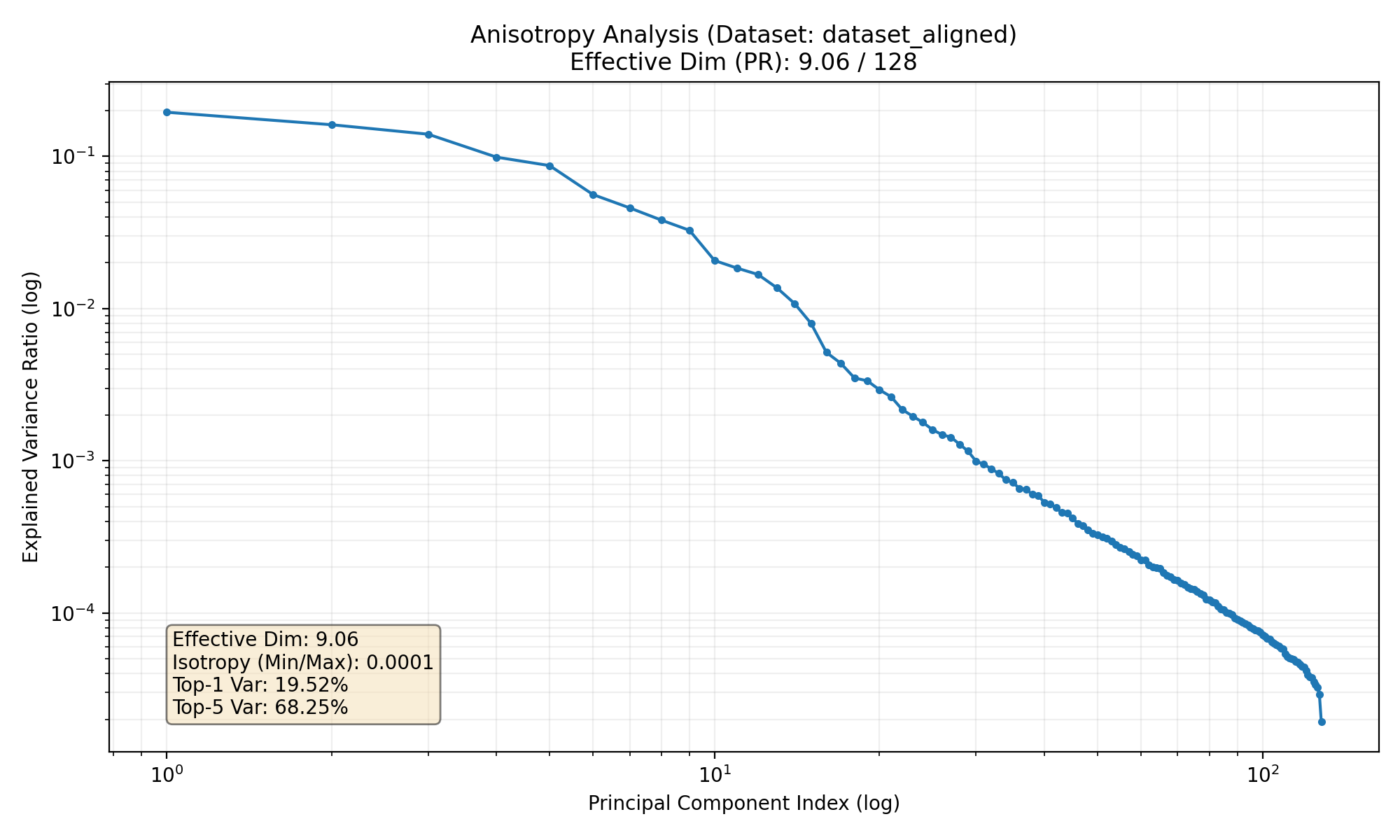}
\caption{Eigenvalue spectrum analysis showing the isotropy characteristics of learned embeddings. The distribution reveals the effective dimensionality and structure of the representation space.} \label{fig:isotropy}
\end{figure}

 We analyze the structural and semantic quality of the learned embeddings using several metrics:

\paragraph{Isotropy and Dimensionality.} LEON embeddings reside on a compact manifold, evidenced by an isotropy ratio ($\lambda_{\min}/\lambda_{\max}$) of $0.000099$ and an effective dimension (participation ratio) of only 9.06 out of 128 (Fig.~\ref{fig:isotropy}). This high anisotropy indicates the model prioritizes dominant spatial factors over noise-like uniformity.

\paragraph{Uniformity Analysis.} We evaluate the embedding distribution on the hypersphere using the uniformity metric~\cite{wang2020understanding} $L_{\text{uniform}}(f; t) = \log \mathbb{E}_{x,y} [ e^{-t\|f(x) - f(y)\|^2} ]$. As shown in Table~\ref{tab:uniformity}, at lower temperatures ($t=2$), the value of {$-2.10$} confirms global clustering and high anisotropy. The sharp decrease at higher temperatures ($t=5, 10$) indicates strong local separability, proving LEON groups spatially related regions while maintaining granularity to distinguish individual locations.

\paragraph{Scale Similarity Analysis.} We analyze how well the embeddings preserve spatial similarity across different H3 resolutions. Table~\ref{tab:similarity} shows correlation coefficients between embeddings at different spatial scales. High similarity scores between adjacent scales (e.g., 0.996 mean similarity for 8-9) demonstrate that the embeddings maintain consistent spatial relationships across resolutions.

\begin{figure}%[htbp]
\centering
\includegraphics[width=1\textwidth]{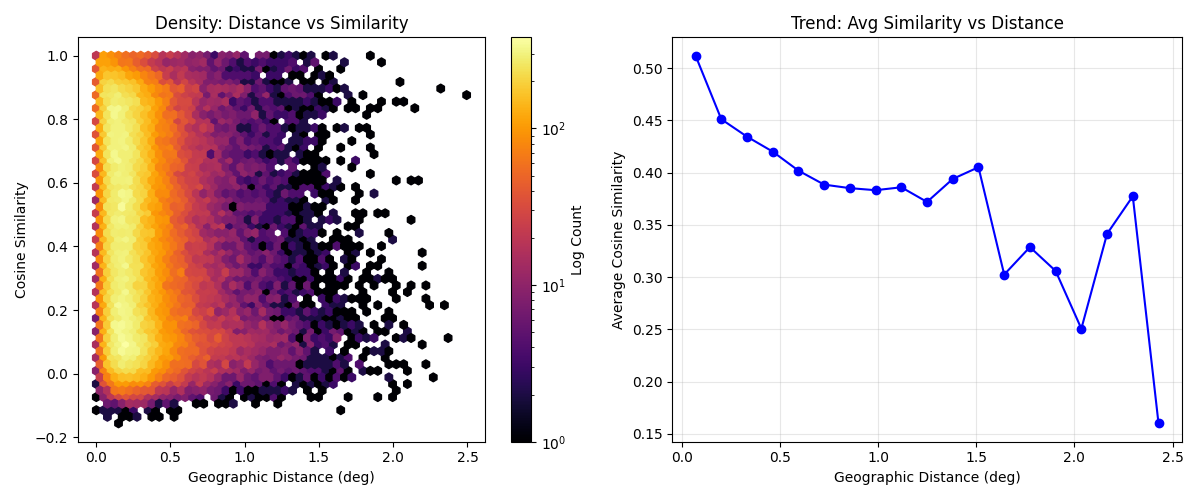}
\caption{Correlation between geographic distance and embedding similarity. The negative correlation ({$-0.116$} Pearson, {$-0.134$} Spearman) demonstrates that spatially proximate locations have more similar representations, confirming spatial autocorrelation in learned embeddings.} \label{fig:correlation}
\end{figure}

\paragraph{Geographic Correlation Analysis.} We evaluate the relationship between geographic distance and embedding similarity, yielding Pearson and Spearman correlations of {$-0.116$} and {$-0.134$}, respectively (Fig.~\ref{fig:correlation}). While the negative correlation validates spatial autocorrelation (local neighbors share features), the low magnitude confirms that embeddings are not merely proxies for coordinates. Instead, the model prioritizes functional semantic similarity over Euclidean proximity; for instance, distant industrial zones show higher similarity than adjacent industrial and residential areas. This demonstrates that LEON encodes land usage rather than raw geographic location.

\begin{figure}[b!]
    \centering
    \begin{subfigure}[htbp]{0.49\textwidth}
        \centering
        \includegraphics[width=\textwidth]{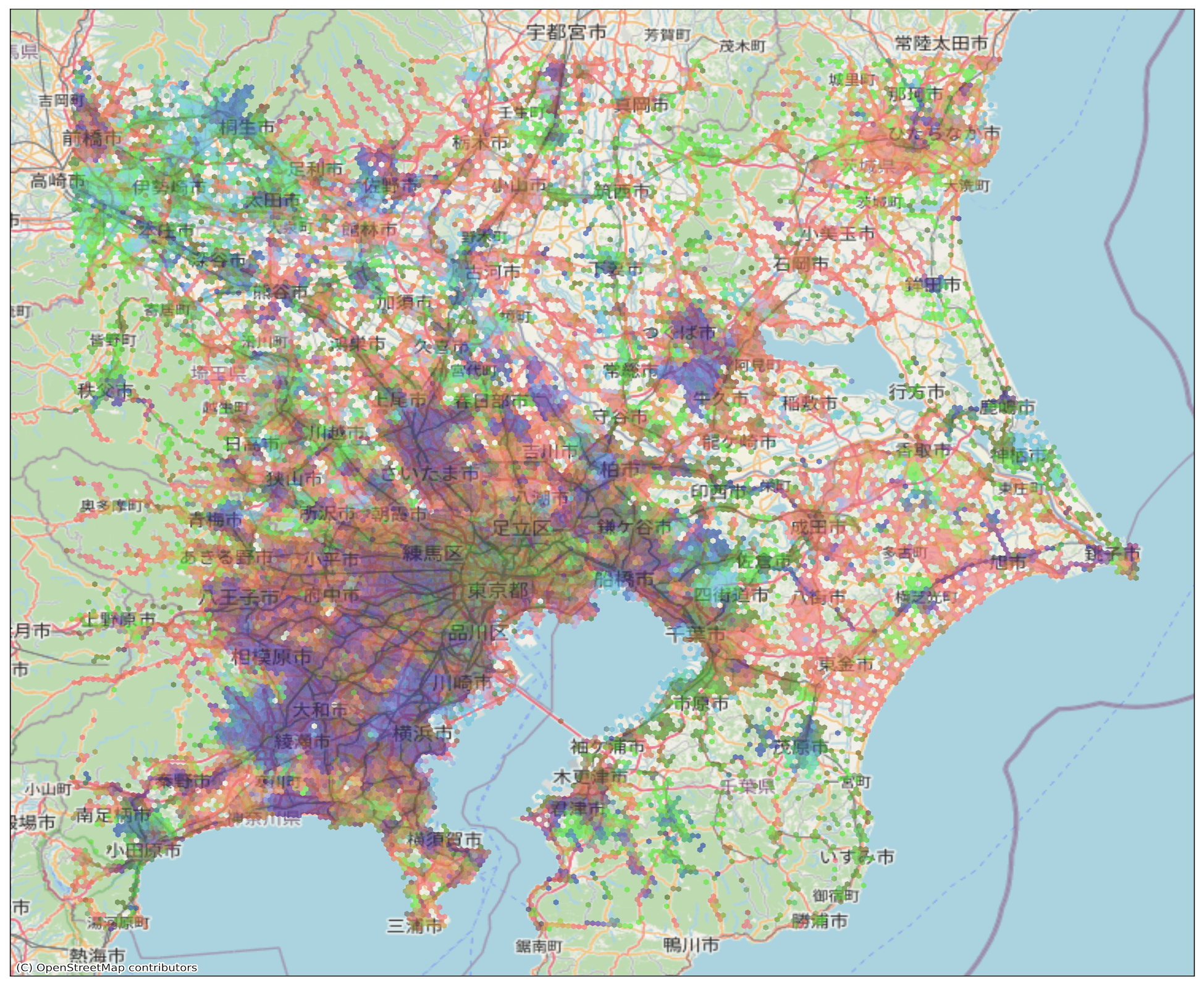}
        \caption{Tokyo: High-density urban corridors}
        \label{fig:map_tokyo}
    \end{subfigure}
    \hfill
    \begin{subfigure}[htbp]{0.49\textwidth}
        \centering
        \includegraphics[width=\textwidth]{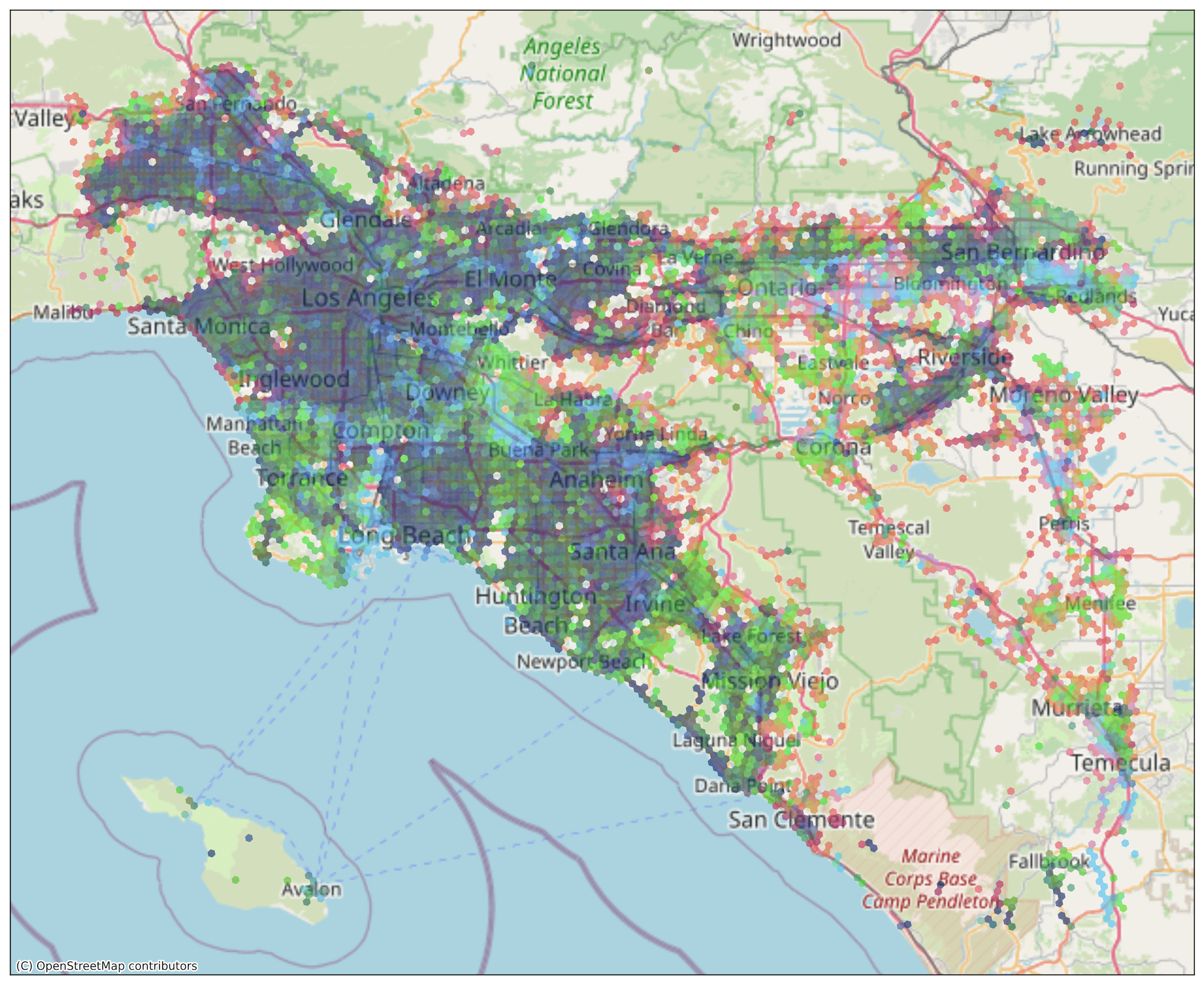}
        \caption{Los Angeles: Terrain-limited sprawl}
        \label{fig:map_la}
    \end{subfigure}
    \vskip\baselineskip
    \begin{subfigure}[htbp]{0.49\textwidth}
        \centering
        \includegraphics[width=\textwidth]{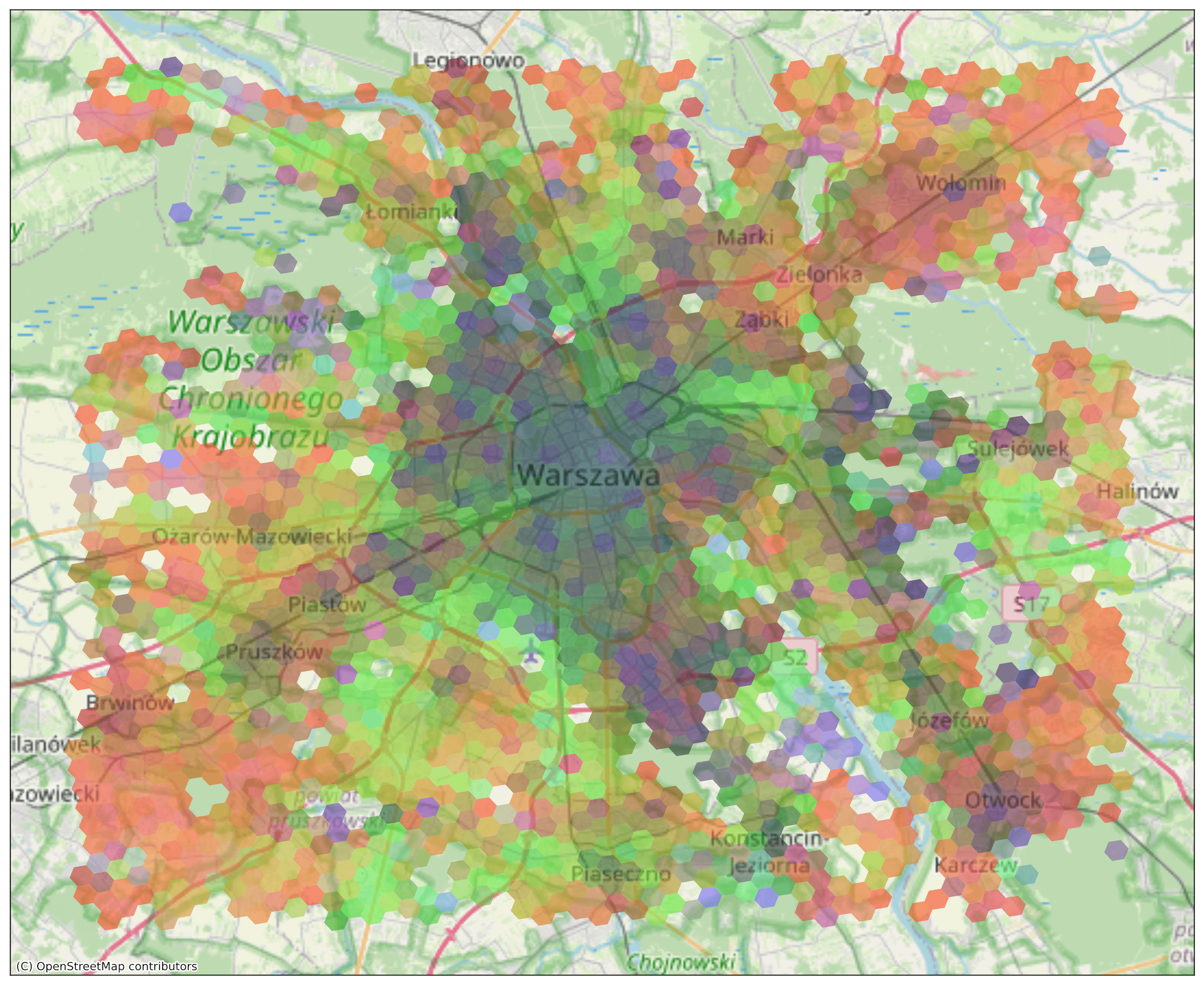}
        \caption{Warsaw: Radial monocentric structure}
        \label{fig:map_warsaw}
    \end{subfigure}
    \hfill
    \begin{subfigure}[htbp]{0.49\textwidth}
        \centering
        \includegraphics[width=\textwidth]{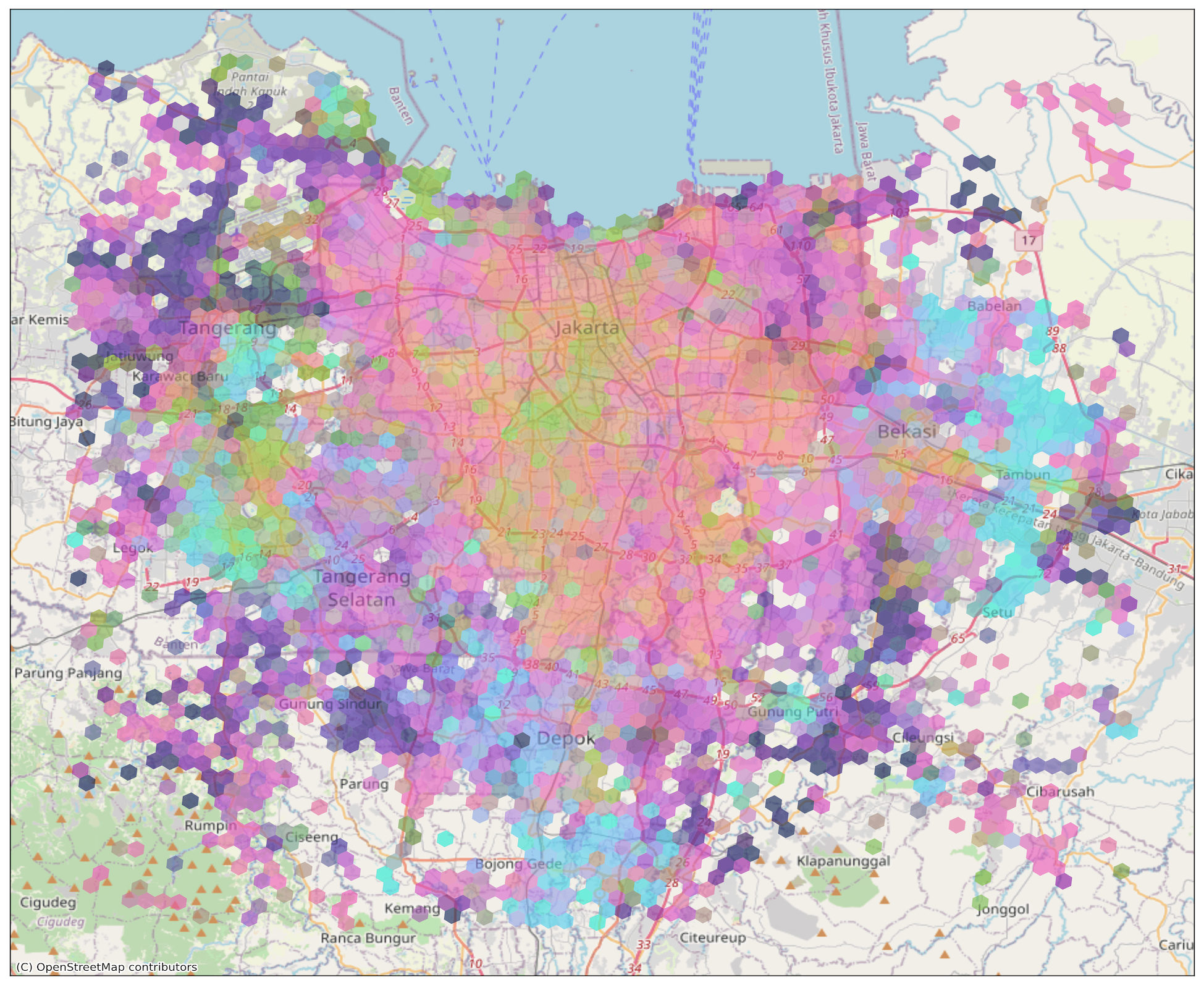}
        \caption{Jakarta: Polycentric urban clusters}
        \label{fig:map_jakarta}
    \end{subfigure}
    \caption{Unsupervised Semantic Mapping. 3D PCA components of embeddings are visualized as RGB colors. The resulting patterns reveal spontaneously discovered urban structures, including transit corridors and industrial zones, where color similarity denotes semantic similarity. }
    \label{fig:semantic_maps}
\end{figure}

\paragraph{Visualization Analysis.} To further validate the semantic capabilities of LEON, we visualize the learned latent space directly on geographic coordinates. We perform Principal Component Analysis on the embeddings of specific cities, reducing the dimensionality to {$3$}, and map these components to the RGB color space. Figure~\ref{fig:semantic_maps} illustrates the resulting semantic maps. The emergence of distinct color gradients reveals that LEON captures complex urban structures without supervision. For example, in Tokyo (Fig.~\ref{fig:map_tokyo}), dense transit corridors are clearly distinguished from residential blocks. In Warsaw (Fig.~\ref{fig:map_warsaw}), the embeddings reconstruct the radial structure of the city, differentiating the historic center from suburban rings. Los Angeles (Fig.~\ref{fig:map_la}) shows linear corridors following transportation networks and valleys.

\begin{figure}%[t!]
\centering
\includegraphics[width=1.0\textwidth]{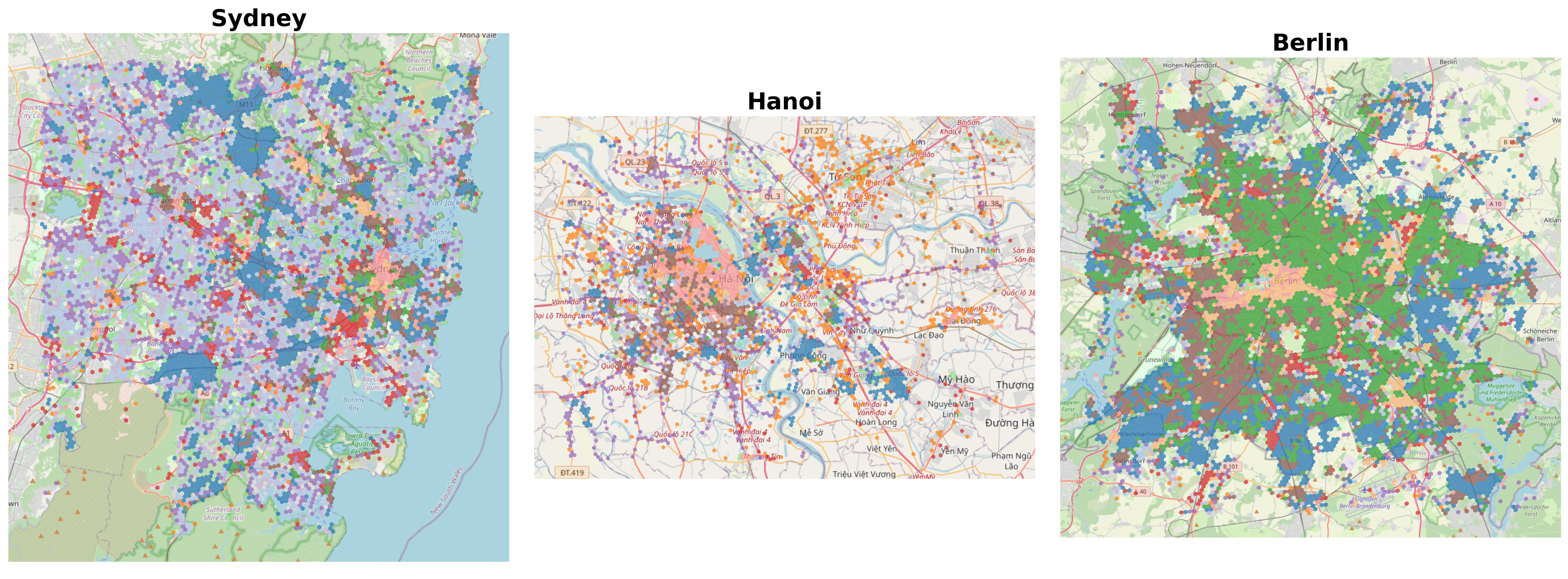}
\caption{Spatial Projection of Universal Semantic Clusters. Embeddings for diverse urban morphologies are partitioned into $K=12$ functional zones. Mapping uses a shared cluster assignment to demonstrate semantic alignment across geographies.}
\label{fig:universal_clusters}
\end{figure}

\paragraph{Semantic Clustering and Functional Zones.} To validate the semantic coherence of the learned latent space, we aggregated embeddings from all {$35$} test cities and performed K-Means clustering \mbox{($K=12$)}. This forces the model to identify universal urban structures across diverse contexts. We interpret clusters using a simple frequency ratio: the mean count of an amenity in a cluster divided by its global mean. A ratio {$>1.0$} indicates the feature characterizes the cluster. The discovered clusters correspond to distinct functional zones, such as Central Business Districts and Residential areas. Figure~\ref{fig:universal_clusters} shows the spatial projection of these clusters. The consistent assignment across cities demonstrates that LEON does not merely memorize local geography, but successfully discovers recurring, universal urban patterns---such as central business districts or sparse residential peripheries---despite completely different urban morphologies.

\begin{table}%[htbp]
\centering
\caption{Uniformity Analysis across different temperature parameters ($t$). Higher temperatures probe local separability, while lower temperatures reflect global distribution.}
\label{tab:uniformity}
\begin{tabular}{cc}
\toprule
Temperature ($t$) & Uniformity Score \\
\midrule
0.5 & -0.6218 \\
1.0 & -1.1799 \\
2.0 & -2.1036 \\
5.0 & -3.7312 \\
10.0 & -4.8699 \\
\bottomrule
\end{tabular}
\end{table}
\begin{table}%[htbp]
\caption{Scale Similarity Analysis}
\label{tab:similarity}
\centering
\begin{tabular}{lccccc}
\toprule
Scale Pair & N & Mean & Std Dev & Min & Max \\
\midrule
7-8 & 900 & 0.9695 & 0.0579 & 0.4940 & 1.0000 \\
7-9 & 900 & 0.9666 & 0.0621 & 0.4786 & 1.0000 \\
8-9 & 900 & 0.9955 & 0.0136 & 0.8766 & 1.0000 \\
Random & 899 & 0.5471 & 0.2774 & -0.0594 & 1.0000 \\
\bottomrule
\end{tabular}
\end{table}

% -----------------------------------------------------------------------------
% 6. DISCUSSION
% -----------------------------------------------------------------------------
\section{Summary, Discussion and Conclusion}

Our results demonstrate several key advantages of the LEON approach: (1) \textbf{Self-supervised Learning} --- LEON does not require labeled data for pre-training, making it applicable to regions with limited ground truth information; (2) \textbf{Scalability} --- the H3-based graph construction scales efficiently to large geographic regions; (3) \textbf{Multi-scale Awareness} --- the embeddings naturally capture patterns at multiple spatial scales; and (4) \textbf{Transferability} --- pre-trained embeddings can be applied broadly across various downstream tasks.
\par This paper introduced LEON, a novel self-supervised framework for learning geospatial region representations using Graph Masked Autoencoders adapted for geographic data. Through comprehensive evaluation on multiple real-world datasets, we demonstrated that LEON can learn meaningful spatial embeddings that improve performance on various geographic prediction tasks.
\par Key contributions include the adaptation of masked autoencoding to spatial graphs, comprehensive analysis of learned embedding properties, and demonstration of scale-invariant spatial representations. The results show that self-supervised learning provides a viable approach for spatial representation learning, particularly when labeled data is scarce.
\par Future work will focus on incorporating temporal dynamics, improving computational efficiency, and developing more sophisticated spatial masking strategies. The integration of multi-modal geographic data and extension to dynamic spatial systems represent promising research directions.
\par Our work contributes to the growing field of spatial representation learning and demonstrates the potential of self-supervised approaches for geographic applications. The learned embeddings and analysis techniques provide a foundation for future research in computational geography and spatial data science.
\subsection{Limitations and Future Work}
\par Several limitations suggest directions for future research:
\begin{enumerate}
    \item \textbf{Scale Adaptation and Multi-Resolution Alignment}: The performance drop observed on the resolution-10 San Francisco dataset underscores the model's sensitivity to out-of-scope spatial shifts. Future work will investigate multi-scale joint pre-training and adaptive grid layers to bridge resolution mismatches explicitly.
    \item \textbf{Structural Trade-offs in Hexagonal Indexing}: While LEON demonstrates notable predictive advantages over traditional baselines and fixed-geometry alternatives like GeoVeX on irregular tasks like Airbnb pricing, a systematic study comparing graph message-passing mechanics against strict convolutional spatial biases is warranted to map task-specific structural preferences.
    \item \textbf{Dynamic and Crowdsourced Artifacts}: Future extensions will attempt to incorporate temporal dynamics into the autoencoding loss and explicitly formulate adjustments for mapping density variances inherent to collaborative OSM data.
\end{enumerate}
\begin{credits}
\subsubsection{\ackname}
We acknowledge the use of publicly available datasets including EuroSAT, Airbnb listings data, and crime statistics from various city governments. We thank the H3 development team at Uber for providing the hierarchical spatial indexing system.

\subsubsection{\discintname}
The authors have no competing interests to declare that are relevant to the content of this article.
\end{credits}

\bibliographystyle{splncs04}
\bibliography{references}

\end{document}